\documentclass[a4paper]{article}
\usepackage{dx}
\usepackage[T1]{fontenc}
\usepackage{ae,aecompl}
\usepackage{amsfonts}
\usepackage{amsmath}
\usepackage{amsthm}

\usepackage{times}

\usepackage[inline]{enumitem}

\usepackage{booktabs}
\usepackage{url}
\usepackage{mathnotation}
\usepackage{ensuretext}
\usepackage{multirow}
\usepackage{booktabs}

\usepackage{balance}

\newcommand{\dpi}{\mathit{dpi}}

\newcommand{\dt}{\mathcal{D}^*}
\newcommand{\md}{\mathcal{D}}
\newcommand{\mD}{\mathbf{D}}

\newcommand{\mo}{\mathcal{K}}
\newcommand{\mb}{\mathcal{B}}
\newcommand{\tax}{\mathit{ax}}
\newcommand{\Tp}{\mathit{P}}
\newcommand{\Tn}{\mathit{N}}
\newcommand{\tp}{\mathit{p}}
\newcommand{\tn}{\mathit{n}}

\newcommand{\allD}{{\bf{allD}}}
\newcommand{\pr}{\mathit{p}}
\newcommand{\oracle}{\mathsf{orcl}}
\newcommand{\ER}{\mathit{er}}
\newcommand{\HERp}{\hat{\ER}_{m,\mD}^+}
\newcommand{\HERn}{\hat{\ER}_{m,\mD}^-}
\newcommand{\Hpp}{\hat{\pr}_{m,\mD}^+}
\newcommand{\Hpn}{\hat{\pr}_{m,\mD}^-}
\newcommand{\Hmpp}[1]{\hat{\pr}_{#1,\mD}^+}
\newcommand{\Hmpn}[1]{\hat{\pr}_{#1,\mD}^-}
\newcommand{\HERmp}[1]{\hat{\ER}_{#1,\mD}^+}
\newcommand{\HERmn}[1]{\hat{\ER}_{#1,\mD}^-}
\newcommand{\pp}{\pr_{m}^+}
\newcommand{\pn}{\pr_{m}^-}
\newcommand{\ERp}{\ER_{m}^+}
\newcommand{\ERn}{\ER_{m}^-}

\newcommand{\rd}{\mathit{rd}}
\newcommand{\ard}{\mathit{ard}}
\newcommand{\bfs}{\mathit{bf}}
\newcommand{\abf}{\mathit{abf}}
\newcommand{\wf}{\mathit{wf}}
\newcommand{\awf}{\mathit{awf}}

\newcounter{examplecounter}
\newenvironment{example}{
	\refstepcounter{examplecounter}%
	
	\vspace{3pt}
	\noindent\textbf{Example \arabic{examplecounter}}%
	\quad
}{
	
	\vspace{3pt}
}

\newenvironment{myquote}[1]%
{\list{}{\leftmargin=#1\rightmargin=#1}\item[]}%
{\endlist}

\title{Do We Really Sample Right in Model-Based Diagnosis?\thanks{This work was accepted for presentation at the \emph{31st International Workshop on Principles of Diagnosis (DX-2020)}. An extended version of this paper was published in the \emph{Proceedings of the 36th AAAI Conference on Artificial Intelligence 2022 (AAAI-2022)} \protect\cite{rodler2022aaai}.}}
\author%
{%
Patrick Rodler$^1$ \and 
Fatima Elichanova$^2$\\
University of Klagenfurt \\
$^1$patrick.rodler@aau.at\\
$^2$fatimael@edu.aau.at
}

\begin{document}

\maketitle

\begin{abstract}
%
Statistical samples, in order to be representative, have to be
drawn from a population in a random and unbiased way. Nevertheless, it is common practice in the field of model-based
diagnosis to make estimations from (biased) best-first samples. One example is the computation of a few most probable possible fault explanations for a defective system and the use of these to assess which aspect of the system, if measured, would bring the highest information gain.

In this work, we scrutinize whether these statistically not
well-founded conventions, that both diagnosis researchers and practitioners have adhered to for decades, are indeed reasonable. To this end, we empirically analyze various sampling methods that generate fault explanations. We study the representativeness of the produced samples in terms of their estimations about fault explanations and how well they guide diagnostic decisions, and we investigate the impact of sample size, the optimal trade-off between sampling efficiency and effectivity, and how approximate sampling techniques compare to exact ones.
\end{abstract}

\section{Introduction}
Suppose we intend to predict the outcome of the next election and we conduct a poll where we ask only, say, university 
professors for whom they are going to vote. By this strategy, we will most likely not gain insight into the real sentiment in the population regarding the election. The problem is 
simply that 
professors are most probably not representative of all people. In model-based diagnosis, however, such kind of samples are often used as a basis for making decisions that rule the efficiency of the diagnostic process.   

\emph{Model-based diagnosis} \cite{Reiter87,dekleer1987} deals with the detection, localization and repair of faults in observed systems such as programs, circuits, knowledge bases or physical devices of various kinds. One important prerequisite to achieve these goals is the generation of 
\emph{diagnoses}, i.e., explanations for the faulty system behavior in terms of potentially faulty system components. A sample of diagnoses can be either
\begin{itemize}[noitemsep]
	\item \emph{directly analyzed}, e.g., to manually discover or make estimations about the actual fault \cite{rodler2019KBS_userstudy,stern2013finding}, 
	to aid proper algorithm choice \cite{jannach2015parallelized,slaney2014set}, or to support users in test case specifications or repair actions \cite{Kalyanpur2006a,DBLP:conf/foiks/SchekotihinRS18,Rodler2015phd,meilicke2011thesis,DBLP:conf/icbo/SchekotihinRSHT18}, or
	\item \emph{used as an input or guidance to diagnostic algorithms},
\end{itemize}
where we focus on the second bullet in this work.

An important class of diagnostic algorithms that are guided by a set of precomputed diagnoses are \emph{sequential diagnosis (SD)} approaches \cite{dekleer1987,dekleer1993}. They use a sample of diagnoses to compute optimal system measurements that allow to efficiently and systematically rule out invalid diagnoses until a single or highly probable one remains. 
Since achieving (global) optimality of the sequence of system measurements is intractable in general \cite{pattipati1990}, state-of-the-art SD techniques usually rely on local optimization \cite{dekleer1992onesteplook} using one out of numerous \emph{heuristics} \cite{dekleer1987,moret1982decision,Shchekotykhin2012,rodler17dx_activelearning,rodler2018ruleML} as optimality criteria.
These heuristics can be seen as functions that, based on a given sample of diagnoses, map measurement candidates to one numeric 
score each, and finally select the one measurement with the best score. In most cases, these functions use two features of the sample:\footnote{For this work, we make the assumption that measurements have uniform costs. If that is not the case, then the measurement cost is another factor that flows into the assessment of measurement candidates \cite{moret1982decision,gonzalez2011spectrum}.} 
\begin{description}[font=\normalfont\em,noitemsep]
	\item[(F1)] the \emph{diagnoses' probabilities} (to estimate the probability of each measurement outcome), and
	\item[(F2)] the \emph{diagnoses' predictions of the measurement outcome} (to estimate diagnosis elimination rates).
\end{description}

Literature offers a wide range of different methods and ways to generate samples of diagnoses, among them ones that return a \emph{specific sample} (which includes exactly a predefined subset of all diagnoses) and others that compute
an \emph{unspecific sample}, e.g., in a heuristic \cite{abreu2009low}, stochastic \cite{feldman2008computing} or simply undefined way \cite{Shchekotykhin2014}\footnote{Note, the algorithm described in \cite{Shchekotykhin2014} can be modified for heuristic diagnosis computation, as will be explained in Sec.~\ref{sec:sample_types}.} 
 (where no guarantee 
 can be given regarding diagnosis selection for the sample).

Many existing SD approaches draw on samples of the specific type in that they build upon \emph{best-first} samples, 
such as maximum-probability or minimum-cardinality diagnoses \cite{Rodler2015phd,Shchekotykhin2012,gonzalez2011spectrum,dekleer1991focusing_prob_diag,dekleer1989behavioral_modes,dekleer1995trading,zamir2014}. 
While perhaps often being motivated by the desideratum that the most preferred/likely candidate(s) should be known at any stage of the diagnostic process, e.g., to allow for well-founded stopping criteria, the use of such non-random samples is highly questionable from the statistical viewpoint.    

In this work we challenge the validity of the following 
statistical law 
in the domain of 
model-based diagnosis: 
\begin{myquote}{10pt}
	\emph{A randomly chosen unbiased sample from a population allows (on average) better conclusions and estimations about the whole population than any other sample.}
\end{myquote}
The motivation behind this inquiry is to either understand why this fundamental principle does not apply to the particular domain of model-based diagnosis, or to rationalize the necessity of random sampling as part of diagnostic algorithms and to foster research in this direction.   

The particular contributions are: 
\begin{itemize}[noitemsep]
	\item We analyze a range of real-world diagnosis problems and gain insight into the quality of \emph{three specific} (best-first, random and, as a baseline, worst-first) \emph{and three unspecific} (approximate best-first, approximate random, approximate worst-first) \emph{sample types}.
	\item We assess a sample type's quality based on 
	\begin{itemize}[noitemsep,topsep=0pt]
		\item its \emph{theoretical representativeness}, i.e., how well it allows to estimate the aspects (F1 and F2) that determine the heuristic score of measurements, and
		\item its \emph{practical representativeness}, i.e., its performance achieved in a diagnosis session wrt.\ time and number of measurements.
	\end{itemize}
	\item We investigate the \emph{impact of the} 
	\begin{itemize}[noitemsep,topsep=0pt]
		\item \emph{sample size},
		\item particular used \emph{heuristic}, and
		\item tackled \emph{diagnosis problem}
	\end{itemize} 
	on the sample's 
	representativeness. 
\end{itemize}

This work is organized as follows: We provide a brief account of theoretical foundations in Sec.~\ref{sec:basics}. Our evaluations (dataset, sample types, sampling techniques, evaluation criteria, research questions, experiment settings, and results) are discussed in Sec.~\ref{sec:eval}. 
In Sec.~\ref{sec:research_limitations}, we address limitations of our research, and conclude with Sec.~\ref{sec:conclusion}. 

\section{Preliminaries}
\label{sec:basics}
We briefly characterize 
concepts from model-based diagnosis used in this work, based on the framework of \cite{Shchekotykhin2012,Rodler2015phd} which is 
more general \cite{rodler17dx_reducing} than Reiter's theory \cite{Reiter87}.\footnote{The main reason for using this more general framework is its ability to handle negative measurements (things that must \emph{not} be true for the diagnosed system)
	which are helpful, e.g., for diagnosing knowledge bases \cite{DBLP:journals/ai/FelfernigFJS04,Shchekotykhin2012}.}

\noindent\textbf{Diagnosis Problem.}
We assume that the diagnosed system, consisting of a set of components $\setof{c_1,\dots,c_k}$, is described by a finite set of logical sentences $\mo \cup \mb$, where $\mo$ (possibly faulty sentences) includes knowledge about the behavior of the system components, and $\mb$ (correct background knowledge) comprises any additional available system knowledge and system observations. More precisely, there is a one-to-one relationship between sentences $\tax_i \in \mo$ and components $c_i$, where $\tax_i$ describes the nominal behavior of $c_i$ (\emph{weak fault model}). E.g., if $c_i$ is an AND-gate in a circuit, then $\tax_i := out(c_i) = and(in1(c_i),in2(c_i))$; $\mb$ in this case might contain sentences stating, e.g., which components are connected by wires, or observed circuit outputs. 
The inclusion of a sentence $\tax_i$ in $\mo$ corresponds to the (implicit) assumption that $c_i$ is healthy. Evidence about the system behavior is captured by sets of positive ($\Tp$) and negative ($\Tn$) measurements \cite{Reiter87,dekleer1987,DBLP:journals/ai/FelfernigFJS04}. Each measurement is a logical sentence; positive ones $\tp\in\Tp$ must be true and negative ones $\tn\in\Tn$ must not be true. The former can be, depending on the context, e.g., observations about the system, probes or required system properties. 
The latter model properties that must not hold for the system, e.g., if $\mo$ is a biological knowledge base to be debugged, a negative test case might be ``every bird can fly'' (think of penguins).
We call $\tuple{\mo,\mb,\Tp,\Tn}$ a \emph{diagnosis problem instance (DPI)}. 

\begin{example} \hspace{-1em}\emph{(DPI)}\quad\label{ex:dpi}
	Table~\ref{tab:example_DPI} depicts a DPI stated in propositional logic. 
	The ``system'' (the knowledge base itself in this case) comprises five ``components'' $c_1, \dots,c_5$, and the ``normal behavior'' of $c_i$ is given by the respective axiom $\tax_i \in \mo$. No background knowledge ($\mb = \emptyset$) or positive measurements ($\Tp=\emptyset$) are given from the start. But, there is one negative measurement (i.e., $\Tn = \setof{\lnot A}$), which stipulates that $\lnot A$ must \emph{not} be an entailment of the correct system (knowledge base). Note, however, that $\mo$ (i.e., the assumption that all ``components'' are normal) in this case does entail $\lnot A$ (e.g., due to the axioms $\tax_1,\tax_2$) and therefore some axiom (``component'') in $\mo$ must be faulty.\qed
\end{example}

\begin{table}
	\centering
	\scriptsize
	\renewcommand{\arraystretch}{1}
	\begin{tabular}{@{}ccc@{}}
		\toprule
		\multicolumn{1}{ c  }{\multirow{2}{*}{$\mo\;=$} } & \multicolumn{2}{ l  }{$\{ \tax_1: A \to \lnot B$ \;\; $\tax_2: A \to B$ \;\; $\tax_3: A \to \lnot C$} \\
		& \multicolumn{2}{ l  }{$\phantom{\{} \tax_4: B \to C$ \;\;\,\hspace{4pt} $\tax_5: A \to B \lor C \qquad\qquad\,\quad\;\}$} \\
		\cmidrule{1-3}
		\multicolumn{3}{ l  }{$\mb =\emptyset \quad\qquad\qquad\qquad \Tp=\emptyset \quad\qquad\qquad\qquad \Tn=\setof{\lnot A}$} \\
		\bottomrule
	\end{tabular}
	\caption{\small Example DPI stated in propositional logic.}
	\label{tab:example_DPI}
\end{table}

\noindent\textbf{Diagnoses.}
Given that the system description along with the positive measurements (under the 
assumption $\mo$ that all components are healthy) is inconsistent, i.e., $\mo \cup \mb \cup \Tp \models \bot$, or some negative measurement is entailed, i.e., $\mo \cup \mb \cup \Tp \models \tn$ for some $\tn \in \Tn$, some assumption(s) about the healthiness of components, i.e., some sentences in $\mo$, must be retracted. We call such a set of sentences $\md \subseteq \mo$ a \emph{diagnosis} for the DPI $\tuple{\mo,\mb,\Tp,\Tn}$ iff $(\mo \setminus \md) \cup \mb \cup \Tp \not\models x$ for all $x \in \Tn \cup \setof{\bot}$. We say that $\md$ is a \emph{minimal diagnosis} for $\dpi$ iff there is no diagnosis $\md' \subset \md$ for $\dpi$. The set of minimal diagnoses is representative of all diagnoses under the weak fault model \cite{Kleer1992}, i.e., 
the set of all diagnoses is equal to 
the set of all supersets of minimal diagnoses.
Therefore, diagnosis approaches usually restrict their focus to only minimal diagnoses. 
We furthermore denote by $\dt$ the \emph{actual diagnosis} which pinpoints the actually faulty axioms, i.e., all elements of $\dt$ are in fact faulty and all elements of $\mo\setminus\dt$ are in fact correct.

\begin{example} \hspace{-1em}\emph{(Diagnoses)}\quad\label{ex:diagnoses}
	For our DPI in Table~\ref{tab:example_DPI} we have four minimal diagnoses, given by $\md_1:=[\tax_1,\tax_3]$, $\md_2:=[\tax_1,\tax_4]$, $\md_3:=[\tax_2,\tax_3]$, and $\md_4 := [\tax_2,\tax_5]$.
	For instance, $\md_1$ is a minimal diagnosis as $(\mo\setminus\md_1) \cup \mb\cup \Tp = \setof{\tax_2,\tax_4,\tax_5}$ is both consistent and does not entail the given negative measurement $\lnot A$.\qed
\end{example}

\noindent\textbf{Diagnosis Probability Model.}
In case useful meta information is available that allows to assess the likeliness of failure for system components, the probability of diagnoses (of being the actual diagnosis) can be derived.
Specifically, given a function $\pr$ that maps each sentence (system component) $\tax \in \mo$ to its failure probability $\pr(\tax) \in (0,1)$, the probability $\pr(\md)$ of a diagnosis 
$\md \subseteq \mo$ (under the common assumption of independent component failure) is computed as the probability that all sentences in $\md$ are faulty, and all others are correct, i.e., 
$
\pr(\md) := \prod_{\tax \in \md} \pr(\tax) \prod_{\tax \in \mo\setminus \md} (1-\pr(\tax)) 
$.
Each time a new measurement is added to the DPI, the probabilities of diagnoses are updated using Bayes' Theorem \cite{dekleer1987}.

\begin{example} \hspace{-1em}\emph{(Diagnosis Probabilities)}\quad\label{ex:diag_probs}
	Reconsider the DPI 
	from
	Table~\ref{tab:example_DPI} and let 
	probabilities $\langle \pr(\tax_1), \dots,$ $\pr(\tax_5)\rangle = \langle .1,$ $.05, .1, .05, .15\rangle$. Then, the probabilities of all minimal diagnoses from Example~\ref{ex:diagnoses} are $\langle \pr(\md_1), \dots,$ $\pr(\md_4)\rangle = \langle .0077,.0036,.0036,.0058\rangle$. E.g., $\pr(\md_1)$ is calculated as $.1*(1-.05)*.1*(1-.05)*(1-.15)$. The normalized diagnosis probabilities would then be $\langle .37,.175,.175,.28\rangle$. Note, this normalization makes sense if only a proper subset of all diagnoses is known.
	\qed 
\end{example}

\noindent\textbf{Measurement Points.}
We call a logical sentence a \emph{measurement point (MP)} if it states one (true or false) aspect of the system under consideration. E.g., if the system is a digital circuit, the statement $out(c_i) = 1$, which states that the output of gate $c_i$ is high, is an MP. In case of the system being, say, a biological knowledge base, $\forall X (bird(X) \to canFly(X))$ is an MP. 
Assuming an oracle $\oracle$ (e.g., an electrical engineer for a circuit, or a domain expert for a knowledge base) that is knowledgeable about the system, one can send to $\oracle$ MPs $m$ and $\oracle$ will classify each $m$ as either a positive or a negative measurement, i.e., $m \mapsto \oracle(m)$ where $\oracle(m) \in \{\Tp,\Tn\}$.    

\noindent\textbf{Measurements to Discriminate between Diagnoses.}
MPs come into play when there are multiple diagnoses for a DPI and the intention is to figure out the actual diagnosis. 
Hence, given a set of diagnoses $\mD$ for a DPI between which we want to discriminate, the MPs $m$ of particular interest are those for which each classification $\oracle(m)$ is inconsistent with some diagnosis in $\mD$ \cite{dekleer1987,Rodler2015phd}. We call such MPs \emph{informative} (wrt.\ $\mD$). In other words, each outcome of a measurement for some informative MP will invalidate some diagnosis.    

Each MP $m$ partitions any set of (minimal) diagnoses $\mD$ into 
subsets $\mD_{m}^+$, $\mD_{m}^-$ and $\mD_{m}^0$ as follows:
\begin{itemize}[noitemsep]
	\item Each $\md\in\mD_{m}^+$ is consistent (only) with $\oracle(m) = \Tp$. (\emph{diagnoses predicting the positive outcome})
	\item Each $\md\in\mD_{m}^-$ is consistent (only) with $\oracle(m) = \Tn$. (\emph{diagnoses predicting the negative outcome})
	\item Each $\md\in\mD_{m}^0$ is consistent with both outcomes $\oracle(m) \in \{\Tp,\Tn\}$.  (\emph{uncommitted diagnoses})	
\end{itemize} 
Thus, an MP $m$ is informative iff both $\mD_{m}^+$ (diagnoses invalidated if $\oracle(m) = \Tn$) and $\mD_{m}^-$ (diagnoses invalidated if $\oracle(m) = \Tp$) are non-empty sets.

\noindent\textbf{(Estimated) Properties of Measurement Points.} Since not all informative MPs are equally utile, the consideration of additional properties of MPs allows a more fine-grained preference rating of MPs. 
In fact, if $\mD$ includes all diagnoses for the given DPI, 
the partition $\langle\mD_{m}^+, \mD_{m}^-, \mD_{m}^0\rangle$ allows to determine, for each measurement outcome $c \in \{\Tp,\Tn\}$, its \emph{diagnosis elimination rate} $\ER(\oracle(m)=c)$ as well as its \emph{probability} $\pr(\oracle(m)=c)$ \cite{rodler17dx_activelearning}: 
\begin{align*}
\ERp := \ER(\oracle(m)=\Tp)&=\tfrac{|\mD_{m}^-|}{|\mD|} \\
\ERn := \ER(\oracle(m)=\Tn)&=\tfrac{|\mD_{m}^+|}{|\mD|} \\
\pp := \pr(\oracle(m)=\Tp)&=P_m^+ + \tfrac{1}{2} P_m^0 \\
\pn := \pr(\oracle(m)=\Tn)&=P_m^- + \tfrac{1}{2} P_m^0 
\end{align*}
where $P_m^X := \sum_{\md\in\mD_{m}^X} \pr(\md)$ for $X \in \{+,-,0\}$.
In practice,
the calculation of 
all diagnoses is often infeasible and diagnosis systems rely on a subset of the minimal diagnoses $\mD$ to \emph{estimate} these properties of MPs. In the following, we denote by 
$\HERp$ and $\HERn$
the \emph{estimated elimination rate} for positive and negative measurement outcome for MP $m$ computed based on $\mD$. Similarly, we refer by 
$\Hpp$ and $\Hpn$
to the \emph{estimated probability} of a positive and negative measurement outcome for $m$ and $\mD$. 
Importantly, these estimated values depend on \emph{both} the MP $m$ \emph{and} the used sample $\mD$ of diagnoses. Note that all four estimates attain values in $[0,1]$ for any MP $m$, and in $(0,1)$ if the MP $m$ is informative. Moreover, $\Hpp + \Hpn = 1$ and $\HERp + \HERn \leq 1$ where the difference $1 - (\HERp + \HERn)$ is the rate 
of uncommitted diagnoses, which are 
not affected by the measurement at $m$.

\begin{example} \hspace{-1em}\emph{(Measurement Points and their Properties)}\quad\label{ex:measurement_points}
	Assume again our DPI from Table~\ref{tab:example_DPI} and let all minimal diagnoses be known, i.e., $\mD = \{\md_1,\dots,\md_4\}$ (cf.\ Example~\ref{ex:diagnoses}). Then, e.g., $m_1 := A \to C$ is an informative MP wrt.\ $\mD$ since $\mD_{m1}^+ = \{\md_1,\md_3\} \neq \emptyset$ and $\mD_{m1}^- = \{\md_2,\md_4\} \neq \emptyset$. E.g., $\md_1 \in \mD_{m1}^+$ holds because $(\mo\setminus\md_1) \cup \mb \cup \Tp = \{\tax_2,\tax_4,\tax_5\} \supset \{A\to B,B\to C\} \models m_1$ and thus $m_1$ can be no negative measurement under the assumption $\md_1$. In a similar way, we obtain that $\md_2 \in \mD_{m1}^-$ due to $(\mo\setminus\md_2) \cup \mb \cup (\Tp \cup \{m_1\}) = \{\tax_2,\tax_3,\tax_5,m_1\} \supset \{A\to \lnot C,A\to C\} \models \lnot A$ where $\lnot A$ is a negative measurement; hence, $m_1$ cannot be a positive measurement under the assumption $\md_2$. In contrast, e.g., $m_2 := B$ is a non-informative MP because $\mD_{m2}^+ = \emptyset$.
	
	Assuming the (normalized) probabilities from Example~\ref{ex:diag_probs}, we obtain probabilities $\Hmpp{m1} = .545, \Hmpn{m1} = .455$ and elimination rates $\HERmp{m1} = .5, \HERmn{m1} = .5$ for $m_1$. Note: \emph{(1)}~If we have at hand a different sample $\mD$, the estimations for one and the same MP might vary substantially. E.g., suppose $\mD = \{\md_1,\md_2,\md_3\}$; then $\Hmpp{m1} = .758, \Hmpn{m1} = .242$ and $\HERmp{m1} \approx 0.33, \HERmn{m1} \approx .67$. \emph{(2)}~If the sample gets smaller (wrt.\ subset-inclusion), then the number of informative MPs might shrink, and vice versa. E.g., $m_1$ becomes non-informative if, e.g., $\mD = \{\md_1,\md_3\}$, and thus might be disregarded by diagnosis systems. Consequently, larger (smaller) samples will tend to provide a richer (sparser) selection of MP candidates.\qed  
\end{example}

\noindent\textbf{Evaluating Measurement Points Using Heuristics.}
To quantitatively assess the preferability of different MPs, 
state-of-the-art sequential diagnosis systems rely on heuristics that 
perform a one-step-lookahead analysis of MPs \cite{dekleer1992onesteplook}. A \emph{heuristic} $h$ is a function that maps each MP $m$ to a real-valued score $h(m)$ \cite{DBLP:journals/corr/Rodler16a}.
At this, $h(m)$ quantifies the \emph{utility of the expected situation after knowing the outcome for MP $m$}. 
The MP 
with the best score 
according to the used heuristic 
is then chosen as a next query to the oracle. 

Well-known heuristics 
incorporate exactly the two discussed features, i.e., the estimated elimination rates 
and estimated probabilities, into 
their computations \cite{rodler17dx_activelearning}. 
So, different heuristics correspond to different functions of these estimates, e.g.: 
\begin{itemize}[noitemsep]
	\item \emph{information gain (ENT)} \cite{dekleer1987} uses solely the probabilities and prefers MPs where $P_m^0 = 0$ and $|\Hpp - \Hpn|$ is minimal \cite{DBLP:journals/corr/Rodler16a};
	\item \emph{split-in-half (SPL)} \cite{Shchekotykhin2012} considers only the elimination rates and favors MPs with $\HERp + \HERn = 1$ and minimal $|\HERp - \HERn|$ \cite{DBLP:journals/corr/Rodler16a};
	\item \emph{risk optimization (RIO)} \cite{Rodler2013} takes into account both features by computing a dynamically re-weighted function of ENT and SPL;
	\item \emph{most probable singleton (MPS)} \cite{rodler17dx_activelearning,DBLP:journals/corr/Rodler16a} also regards both features by giving preference to MPs that maximize the probability of a maximal elimination rate.
\end{itemize}
For details on these and other heuristics see \cite{rodler17dx_activelearning} for a theoretical analysis and \cite{rodler2018ruleML} for an empirical evaluation.
%

\begin{example} \hspace{-1em}\emph{(Heuristics)}\quad\label{ex:heuristics}
	Reconsider our DPI from Table~\ref{tab:example_DPI} and the MP $m_1$ from Example~\ref{ex:measurement_points}, and let all minimal diagnoses be known, i.e., $\mD = \{\md_1,\dots,\md_4\}$. Further, let $m_3 := A\land\lnot B \to C$. Note that $m_3$ is informative (wrt.\ $\mD$), $\mD_{m3}^+ = \{\md_1,\md_2,\md_3\}$, $\mD_{m3}^- = \{\md_4\}$, and the estimations $\Hmpp{m3} = .72, \Hmpn{m3} = .28$ and $\HERmp{m3} = .25, \HERmn{m3} = .75$. Hence, given the two MP candidates $\{m_1,m_3\}$, the heuristic SPL would select $m_1$ (since a half of the known diagnoses are eliminated for each outcome). Similarly, ENT would prefer $m_1$ to $m_3$ (because for $m_1$ roughly a half of the probability mass is eliminated for each outcome). 
	
	However, assume that a used diagnosis sampling technique just outputs the sample $\mD = \{\md_2,\md_3,\md_4\}$. In this case, we obtain the probability estimates $\Hmpp{m1} = .28, \Hmpn{m1} = .72$ and $\Hmpp{m3} = .55, \Hmpn{m3} = .45$, respectively. Hence, using ENT, the chosen MP would be  $m_3$ (the worse MP, as shown above). If sampling would yield $\mD = \{\md_2,\md_4\}$, then $m_1$ would not even be an informative MP (wrt.\ $\mD$) on the one hand, and $m_3$ would be the (theoretically) optimal MP according to SPL on the other hand. This example shows the dramatic impact the used sampling technique can have on diagnostic decisions.\qed  
\end{example}

\noindent\textbf{Sequential Diagnosis (SD)} aims at generating a sequence of informative MPs such that a single (highly probable) diagnosis remains for the given DPI, while minimizing the number of MPs needed (oracle inquiries are usually expensive). A generic SD process iterates through the following steps until (the Bayes-updated) $\pr(\md)$ for some $\md \in \mD$ exceeds a probability threshold $\sigma$:
\begin{enumerate}[noitemsep,label=\emph{S\arabic*}]
	\item \label{SD:step:diagnosis_sampling} Generate a sample of minimal diagnoses $\mD$ for the current DPI (initially, the given DPI).
	\item \label{SD:step:measurement_selection} Choose a (heuristically optimal) informative MP $m$ wrt.\ $\mD$ (using a selection heuristic $h$).
	\item \label{SD:step:oracle_inquiry} Ask the oracle $\oracle$ to classify $m$.
	\item \label{SD:step:knowledge_update} Use the classification $\oracle(m)$ to update the DPI, by adding $m$ to the positive measurements if $\oracle(m)=\Tp$ and to the negative measurements if $\oracle(m)=\Tn$. 
\end{enumerate}

\section{Evaluation}
\label{sec:eval}
We conducted extensive experiments using a dataset of real-world diagnosis cases (Sec.~\ref{sec:dataset}) to study six different diagnosis sample types (Sec.~\ref{sec:sample_types} and \ref{sec:sampling_techniques}) wrt.\ the accuracy of estimations 
and diagnostic efficiency (Sec.~\ref{sec:evaluating_samples}) in different scenarios in terms of sample size (number of diagnoses computed) and measurement selection heuristic used. The concrete research questions are explicated in Sec.~\ref{sec:research_questions} and the experiments are detailed in Sec.~\ref{sec:experiments}. Finally, in Sec.~\ref{sec:results}, we present and discuss the obtained results.  

\subsection{Dataset}
\label{sec:dataset}
In our experiments we drew upon the set of real-world diagnosis problems from the domain of knowledge-base debugging shown in Table~\ref{tab:dataset}. Note that every model-based diagnosis problem (according to Reiter's original characterization \cite{Reiter87}) can be represented as a knowledge-base debugging problem \cite{rodler17dx_reducing}, which is why considering knowledge-base debugging problems is without loss of generality. 
To obtain a representative dataset 
we chose it in a way it covers a variety of different problem sizes, theorem proving complexities, and diagnostic metrics (number of diagnoses, their sizes, number of conflicts, number of components). These metrics are depicted in the columns of Table~\ref{tab:dataset}. 
In order to implement the random sampling of diagnoses, another requirement to the dataset was that all the used 
problems allow the computation of \emph{all} minimal diagnoses within tolerable time for our experiments (single digit number of minutes). 

\setlength{\tabcolsep}{4pt}
\begin{table}
	\renewcommand\arraystretch{1}
	\scriptsize
	\centering
	\caption{\small Dataset used in experiments (sorted by 2nd column).}
	\label{tab:dataset}
	\begin{minipage}{0.98\linewidth}
		\centering
		\begin{tabular}{@{}lrlrr@{}} 
			\toprule
			KB $\mo$				& $|\mo|$& expressivity \textsuperscript{\textbf{1)}} 		& \#D/min/max \textsuperscript{\textbf{2)}} 	& \#C/min/max \textsuperscript{\textbf{2)}} \\ \midrule
			University (U) 
			\textsuperscript{\textbf{3)}}
			& 50 		& $\mathcal{SOIN}^{(D)}$& 90/3/4  &  4/3/5  \\
			IT  
			\textsuperscript{\textbf{4)}}			
			& 
			140 		& $\mathcal{SROIQ}$& 1045/3/7	&  7/3/7  \\
			UNI  
			\textsuperscript{\textbf{4)}}		
			& 
			142 		& $\mathcal{SROIQ}$& 1296/5/6	&  6/3/10 \\
			MiniTambis (M) 
			\textsuperscript{\textbf{3)}}	
			& 173 		& $\mathcal{ALCN}$ 		& 48/3/3  &	 3/2/6 \\
			
			Transportation (T) 
			\textsuperscript{\textbf{3)}}	
			& 
			1300 		& $\mathcal{ALCH}^{(D)}$& 1782/6/9	&  9/2/6  \\
			Economy (E) 
			\textsuperscript{\textbf{3)}}	
			& 1781 		& $\mathcal{ALCH}^{(D)}$& 864/4/8   &  8/3/4 \\
			DBpedia (D) \textsuperscript{\textbf{5)}}			
			& 
			7228 		& $\mathcal{ALCHF}^{(D)}$& 7/1/1  &  1/7/7  \\
			Cton (C) \textsuperscript{\textbf{6)}}			
			& 
			33203		& $\mathcal{SHF}$& 15/1/5   &  6/3/7  \\
			\bottomrule
		\end{tabular}
	\end{minipage}
	\renewcommand\arraystretch{1}
	\begin{minipage}{0.98\linewidth}
		\setlength{\tabcolsep}{2pt}
		\begin{tabular}{@{}lp{7.75cm}@{}}
			\textbf{1):} & Description Logic expressivity
			\cite{DLHandbook}; the higher the expressivity, 
			the higher is the complexity of consistency checking (conflict computation) for this logic.
			\\
			\textbf{2):} & \#D/min/max denotes the number/the minimal size/the maximal size of minimal diagnoses for the DPI resulting from KB $\mo$. Same notation for conflicts.
			\\
			\textbf{3):} & Sufficiently hard diagnosis problems from evaluations in \cite{Shchekotykhin2012}, which were also used, e.g., in \cite{rodler2018ruleML,horridge2009lemmas,ji2014measuring}. \qquad\textbf{4):} Diagnosis problems studied in \cite{rodler2019KBS_userstudy,rodler2020ecai}.
			\\
			\textbf{5):} & Faulty version of DBpedia ontology, see \url{https://bit.ly/2ZO2qYZ}.
			\\
			\textbf{6):} & Diagnosis problem used in scalability tests in \cite{Shchekotykhin2012}. The second scalability problem used in \cite{Shchekotykhin2012} was not included in the dataset since computation of all minimal diagnoses was infeasible (within hours of computation) for it.
		\end{tabular}
	\end{minipage}
\end{table}

\subsection{Sample Types}
\label{sec:sample_types}
We examined the following types of diagnosis samples:
\begin{enumerate}[label=\emph{T\arabic*},noitemsep] 
	\item \label{T5} best-first ($\bfs$)
	\item \label{T1} random 	($\rd$)
	\item \label{T6} worst-first ($\wf$)
	\item \label{T3} approximate best-first ($\abf$)
	\item \label{T2} approximate random ($\ard$)
	\item \label{T4} approximate worst-first ($\awf$)
\end{enumerate}
By ``best-first'' / ``worst-first'', we mean the most / least probable minimal diagnoses. The types \ref{T6} and \ref{T4} serve as baselines. 
We refer to \ref{T5}, \ref{T1} and \ref{T6} as \emph{specific sample types} because we know the properties of the sample (exactly the $k$ best or worst diagnoses, or $k$ unbiased random ones) in advance by employing (expensive) sampling techniques that guarantee these properties. On the other hand, we call \ref{T3}, \ref{T2} and \ref{T4} \emph{unspecific sample types} and adopt (usually less costly) heuristic techniques to provide them.   
In the following,  we denote a sample of type $Ti$ including $k$ minimal diagnoses by $S_{Ti,k}$.

\subsection{Sampling Techniques}
\label{sec:sampling_techniques}
The approaches we used for generating the samples for a given DPI $\dpi = \langle\mo,\mb,\Tp,\Tn\rangle$ were:

	\noindent\textbf{\ref{T5}:} We used uniform-cost HS-Tree \cite[Sec.~4.6]{Rodler2015phd} and stopped it after $k$ diagnoses were computed. Due to the best-first property of the algorithm, these are provenly \cite[Prop.~4.17]{Rodler2015phd} the $k$ diagnoses with the highest probability among all minimal diagnoses.

	\noindent\textbf{\ref{T1}, \ref{T6}:} 
	We generated all\footnote{This is generally intractable \cite{Bylander1991}.
	So, this approach to random sampling is not viable in practice and just used for the purpose of our evaluation. As said in Sec.~\ref{sec:dataset}, we chose our dataset so that
	computation of $\allD$ was feasible 
	within reasonable time.} minimal diagnoses $\allD$ for $\dpi$. For \ref{T1}, we selected $k$ random elements from this set by means of the Java (v1.8) pseudorandom number generator. 
	For \ref{T6}, we picked the $k$ diagnoses with lowest probability.\footnote{Note, the naive approach to generating the least probable diagnoses using a best-first diagnosis computation mechanism (such as a uniform cost version of Reiter's HS-Tree \cite{Rodler2015phd}) and simply taking the reciprocals $p'(\tax) := 1-p(\tax)$ instead of the probabilities $p(\tax)$ for $\tax\in\mo$ (provably) does not work in general.}
	The generation of $\allD$ can be done by any sound and complete diagnosis computation, e.g., HS-Tree \cite{Reiter87}.

%

	\noindent\textbf{\ref{T3}, \ref{T2}, \ref{T4}:} We used Inv-HS-Tree \cite{Shchekotykhin2014} to supply the samples. First, we added all $\tax \in \mo$ to a list $L$. For \ref{T2}, we randomly shuffled $L$. For \ref{T3} and \ref{T4}, we sorted $L$ in descending and ascending order of probability $\pr(\tax)$, respectively.
	Finally, we let plain Inv-HS-Tree operate on this list $L$ to supply a sample of size $k$.
	
	Inv-HS-Tree uses $k$ calls to a diagnosis computation method called Inverse QuickXPlain (Inv-QX) \cite{Felfernig2011,junker04,rodler2020qx}. Each call of Inv-QX returns one well-defined minimal diagnosis $\md_L$ for $\dpi$ 
	based on the strict total order of elements imposed by the sorting 
	of the list $L$. Specifically, $\md_L$ is the minimal diagnosis with highest rank wrt.\ the antilexicographic order $>_{\mathsf{antilex}}$ defined on sublists of $L = [l_1,\dots,l_{|\mo|}]$ \cite{junker04}. At this, for sublists $X,Y$ of $L$, we have $X >_{\mathsf{antilex}} Y$ \emph{($X$ has higher rank wrt.\ $>_{\mathsf{antilex}}$ than $Y$}) iff there is some $k$ such that $X \cap \{l_{k+1},\dots,l_{|\mo|}\} = Y \cap \{l_{k+1},\dots,l_{|\mo|}\}$ (both sublists are equal wrt.\ their lowest ranked elements in $L$) and $l_k \in Y\setminus X$ (the first element that differs between the sublists is in $Y$).
	%
	E.g., if $L$ includes the letters $a,b,\dots,z$ in alphabetic order, then $X >_{\mathsf{antilex}} Y$ for $X = [b,n,r,v]$ and $Y = [a,p,r,v]$ because both lists share $[r,v]$ and, after deleting these two letters from both $X$ and $Y$, the now last element ($p$) of $Y$ is ranked lower in $L$ than the one ($n$) of $X$ (see also \cite[Sec.~3.2.5]{DBLP:journals/corr/Rodler2017}).
	
	That is, in the approximate best-first case (\ref{T3}), the computed diagnosis $\md_L = [d_1,\dots,d_{n-1},d_n]$ has the property that there is no other minimal diagnosis $\md' = [d'_1,\dots,d'_r]$ where $d'_r$ has higher probability than $d_n$, and among all minimal diagnoses that share the last element $d_n$, there is no other minimal diagnosis whose second-last element has a higher probability than $d_{n-1}$, and so forth. If we replace ``higher probability'' with ``lower probability'', we obtain a description of the diagnosis $\md_L$ returned in the approximate worst-first case (\ref{T4}). In the approximate random case (\ref{T2}), we reshuffle $L$ before each call of Inv-QX, thereby trying to simulate a random selection. Note, Inv-HS-Tree guarantees that each Inv-QX call generates a \emph{new} diagnosis by systematically ``blocking'' different elements in $L$ which must not occur in the next diagnosis \cite{Shchekotykhin2014}.
%
%
	
	

\subsection{Evaluating Samples}
\label{sec:evaluating_samples}
We evaluate sample types based on what we call their theoretical and practical representativeness: 

\noindent\textbf{Theoretical Representativeness:} A sample type $Ti$ is the more representative, the better the 
\begin{itemize}[noitemsep]
	\item probability estimates $\langle \Hpp, \Hpn \rangle$ for MPs $m$ match the respective actual values 
	$\langle \pp, \pn\rangle$,
	\item elimination rate estimates $\langle \HERp, \HERn \rangle$ for MPs $m$ match the respective actual values $\langle \ERp,\ERn \rangle$
\end{itemize}
for samples $\mD = S_{Ti,k}$.

\noindent\textbf{Practical Representativeness:} A sampling technique $Ti$ is the more representative, the lower the 
\begin{itemize}[noitemsep]
	\item number of measurements required,
	\item time required for sampling (diagnoses computation)
\end{itemize}
throughout a sequential diagnosis session until the actual diagnosis is isolated from spurious ones, where 
$\mD = S_{Ti,k}$ in each sequential diagnosis iteration. 

\subsection{Research Questions} 
\label{sec:research_questions}
The goal of our evaluation is to shed light on  
the following research questions:
\begin{enumerate}[label=\emph{RQ\arabic*},noitemsep,leftmargin=24pt]
	\item \label{rq1} Which type of sample is best in terms of theoretical representativeness?
	\item \label{rq2} Which type of sample is best in terms of practical representativeness?
	\item \label{rq3} Are the results wrt.\ \ref{rq1} and \ref{rq2} consistent over different \emph{(a)}~sample sizes, \emph{(b)}~measurement selection heuristics, and \emph{(c)}~diagnosis problem instances? 
	\item \label{rq4} Does larger sample size (more computed diagnoses) imply better representativeness?
	\item \label{rq5} Does a better theoretical representativeness translate to a better practical representativeness? 
\end{enumerate}

\subsection{Experiments}
\label{sec:experiments}
We conducted two experiments, EXP1 and EXP2, to investigate our research questions. 
Common to both experiments are the following settings:
\begin{itemize*}[noitemsep]
	\item We defined one DPI $\dpi_\mo := \langle\mo,\emptyset,\emptyset,\emptyset\rangle$ for each $\mo$ in Table~\ref{tab:dataset}. That is, we assumed each axiom (component) in $\mo$ to be possibly faulty and left the background knowledge and the measurements void to begin with. To each $\tax\in\mo$, we randomly assigned a fault probability $\pr(\tax) \in (0,1)$ in a way that syntactically equally (more) complex axioms have an equal (higher) probability (cf.\ \cite{Rodler2015phd,Shchekotykhin2012}). E.g., in our DPI in Table~\ref{tab:example_DPI}, elements of $\{\tax_1,\tax_3\}$ (one implication, one negation) and $\{\tax_2,\tax_4\}$ (one implication), respectively, would each be allocated the same probability, and the former two would have a higher probability than the latter (cf.\ Example~\ref{ex:diag_probs}). 
	\item We precomputed all minimal diagnoses $\allD$ for each DPI $\dpi_\mo$.
	\item We used all sample types $Ti$ for $i\in\{1,\dots,6\}$ (cf.\ Sec.~\ref{sec:sample_types}).
	\item We used sample sizes (numbers of generated minimal diagnoses) $k \in \{2,6,10,20,50\}$.
\end{itemize*}

The specific settings for each experiment were:

\noindent\textbf{EXP1: (theoretical representativeness)}
For each $\dpi_\mo$, for each $k$, and for each $Ti$, we computed a sample $\mD = S_{Ti,k}$. We used 
\begin{itemize}[noitemsep]
	\item $\mD$ to compute probability and elimination rate estimates $\langle \Hpp, \Hpn\rangle$ and $\langle \HERp, \HERn \rangle$, and
	\item $\allD$ to compute $\langle \pp, \pn\rangle$ and $\langle \ERp, \ERn \rangle$
\end{itemize}
for $50$ (if so many, otherwise for all) randomly selected informative MPs wrt.\ $\mD$. For each such MP, we thus had four estimates and four corresponding actual values, that we could compare against one another. 

\noindent\textbf{EXP2: (practical representativeness)} 
For each $\dpi_\mo$, for each $k$, for each $Ti$, and for each of the four heuristics $h \in \{\text{ENT,SPL,RIO,MPS}\}$ (cf.\ Sec.~\ref{sec:basics}), we executed 10 sequential diagnosis sessions (loop \ref{SD:step:diagnosis_sampling}--\ref{SD:step:knowledge_update}, Sec.~\ref{sec:basics}) while in each session
\begin{itemize}[noitemsep]
	\item searching for a different randomly selected target diagnosis $\md^*\in\allD$ for $\dpi_\mo$,
	\item starting from the initial problem $\dpi_\mo$
	\item with stop criterion $\sigma = 1$ (loop until a single minimal diagnosis remains, i.e., all others have been ruled out),
\end{itemize} 
where in each iteration through the loop at step
\begin{itemize}[noitemsep]
	\item \ref{SD:step:diagnosis_sampling}, a sample $\mD = S_{Ti,k}$ is drawn for the current DPI,
	\item \ref{SD:step:measurement_selection}, an informative MP that is optimal for $h$ is selected,
	\item \ref{SD:step:oracle_inquiry}, an automated oracle classifies each MP in a way the predefined target diagnosis $\md^*$ is not ruled out.  
\end{itemize}
For our analyses, we recorded (sampling) times and number of measurements (i.e., loop iterations) throughout a session.

\begin{table}[]
	\scriptsize
	\centering
	\caption{\small Theoretical and practical representativeness: Rankings of sample types for various scenarios (EXP1 and EXP2).}
	\label{tab:rankings}
	\begin{tabular}{@{}l|cccccc|c@{}}
		\toprule
		\textbf{scenario} & \tiny{(best)} &\multicolumn{4}{c}{\textbf{ranking}} & \tiny{(worst)}                      & \textbf{criterion} \\ \midrule
		all                            &  $\rd$ & $\wf$ & $\bfs$ & $\awf$ & ($\abf$ & $\ard$)               &E                  \\
		$k=6$                              &  $\bfs$ &   $\wf$ &   ($\rd$ &   $\awf$) &  $\abf$ &    $\ard$     &  E                  \\
		$k=10$                             &  $\rd$ &   $\wf$ &   $\bfs$ &   $\awf$ &  $\abf$ &    $\ard$       &  E                  \\
		$k=20$                             &  $\rd$ &   $\wf$ &   $\bfs$ &   $\awf$ &  ($\abf$ &    $\ard$)     &  E                  \\
		$k=50$                             &  $\rd$ &   $\wf$ &   $\awf$ &  $\bfs$ &   $\ard$ &  $\abf$         &  E                  \\ \midrule
		all                            &  $\bfs$ &   $\rd$ &   $\awf$ &  $\abf$ &    $\ard$ &  $\wf$        &  P                  \\
		$k=6$                              &  $\bfs$ &   ($\abf$ &  $\rd$ &   $\ard$) & $\awf$ &    $\wf$      &  P                  \\
		$k=10$                             &  $\bfs$ &   $\rd$ &   ($\abf$ &  $\awf$) &    ($\ard$ &  $\wf$)    &  P                  \\
		$k=20$                             &  $\bfs$ &   $\rd$ &   $\awf$ &  ($\abf$ &    $\ard$ &  $\wf$)      &  P                  \\
		$k=50$                             &  $\bfs$ &   $\rd$ &   $\awf$ &  $\ard$ &    $\wf$ &   $\abf$       &  P                  \\ 
		\midrule \midrule
		all                            &  $\bfs$ &   $\ard$ &  $\abf$ &    $\rd$ &   $\awf$ &    $\wf$      &  M                  \\
		$k=2$                             &  $\bfs$ &   $\abf$ &  $\ard$ &    $\awf$ &  $\rd$ &     $\wf$      &  M                  \\
		$k=6$         			           &  $\bfs$ &   $\rd$ &   $\ard$ &  $\abf$ &    $\awf$ &  $\wf$        &  M                  \\
		$k=10$           		           &  $\rd$ &   $\abf$ &  $\ard$ &    $\bfs$ &   $\awf$ &    $\wf$      &  M                  \\
		$k=20$          		           &  $\ard$ &    $\abf$ &  $\awf$ &  $\rd$ &   $\bfs$ &   $\wf$        &  M                  \\
	    $k=50$          		           &  $\ard$ &    $\rd$ &   $\awf$ &    $\bfs$ &   $\abf$ &    $\wf$    &  M                  \\
	    $h=\text{ENT}$                     &  $\bfs$ &   $\abf$ &  $\ard$ &    $\awf$ &  $\rd$ &     $\wf$      &  M                  \\
		$h=\text{SPL}$                     &  $\bfs$ &   $\ard$ &  $\abf$ &    $\rd$ &   $\awf$ &    $\wf$      &  M                  \\
		$h=\text{RIO}$                     &  $\rd$ &   $\ard$ &  $\awf$ &    $\bfs$ &   $\abf$ &    $\wf$     &  M                  \\
		$h=\text{MPS}$                     &  $\ard$ &    $\abf$ &  $\rd$ &     $\awf$ &  $\wf$ &     $\bfs$    &  M                  \\
		 \midrule
		all                            &  $\awf$ &    $\bfs$ &   ($\abf$ &    $\ard$) & $\rd$ &     $\wf$  &  T                  \\
		$k=2$                              &  $\abf$ &    ($\ard$ &  $\awf$) &  $\bfs$ &   $\rd$ &   $\wf$      &  T                  \\
		$k=6$                              &  $\awf$ &    ($\abf$ &  $\ard$) & $\bfs$ &   ($\rd$ &   $\wf$)    &  T                  \\
		$k=10$                             &  $\awf$ &    $\abf$ &  ($\ard$ &  $\bfs$) &   $\rd$ &   $\wf$      &  T                  \\
		$k=20$                             &  $\bfs$ &   $\ard$ &  $\awf$ &    $\abf$ &  $\rd$ &     $\wf$      &  T                  \\
		$k=50$                             &  $\bfs$ &   $\wf$ &   ($\awf$ &  $\rd$) &  $\ard$ &  $\abf$      &  T                  \\
		$h=\text{ENT}$                     &  $\awf$ &    $\abf$ &  $\bfs$ &     $\ard$ &  $\rd$ &     $\wf$    &  T                  \\
		$h=\text{SPL}$                     &  ($\abf$ &    $\awf$ &  $\bfs$) &     $\ard$ &  $\rd$ &     $\wf$  &  T                  \\
		$h=\text{RIO}$                     &  $\awf$ &    ($\abf$ &  $\ard$ &  $\bfs$) &   $\rd$ &   $\wf$      &  T                  \\
		$h=\text{MPS}$                     &  $\bfs$ &   $\awf$ &  $\ard$ &    $\abf$ &  $\rd$ &     $\wf$      &  T                  \\ \bottomrule
	\end{tabular}
\end{table}

\subsection{Results}
\label{sec:results}
From our experiments, we obtained two large datasets, with $6*8*5 = 240$ (EXP1) and $6*8*5*4 = 960$ (EXP2) factor combinations, respectively, for the factors sample type (6 levels), diagnosis problem (8), sample size (5), and heuristic (4). 
Due to the high information content of our data and the paper length restrictions, we can only 
provide a very condensed presentation of the results, given in Tables~\ref{tab:rankings} and \ref{tab:best_wrt_seq-diag-time}.
\textbf{Presentation.} 
\emph{Table~\ref{tab:rankings}} shows rankings of the sample types over different subsets of all factor combinations (referred to as \emph{scenarios}; left column of the table). E.g., scenario ``all'' means all 240 (EXP1) / 960 (EXP2) cases aggregated, whereas ``$k=20$'' denotes exactly the $240:5=48$ (EXP1) / $960:5=192$ (EXP2) cases where the sample size was set to 20. Results from EXP1 are depicted in the top part of the table (first ten rows); results from EXP2 in the bottom part. 
A sample type $Ti$ being ranked prior to type $Tj$ (middle column of table) means that $Ti$ was better than $Tj$ in more of the factor combinations of the respective scenario than vice versa. At this, the meaning of ``better'' (\emph{criterion} for comparison; rightmost table column) is 
\begin{itemize}[noitemsep]
	\item a higher Pearson correlation coefficient between estimated and real values for elimination rate (E)
	and, respectively, probability (P)
	estimations (cf.\ \emph{theoretical representativeness} in Sec.~\ref{sec:evaluating_samples}), and
	\item a lower average number of measurements (M) and, respectively, a lower average sample computation time (T) in sequential sessions (cf.\ \emph{practical representativeness} in Sec.~\ref{sec:evaluating_samples}).
\end{itemize}
 Note, the table does \emph{not} give information about \emph{how much better (worse)} one $Ti$ was than another, but only that it was a preferred choice to the other \emph{in more (less) cases} (of a scenario). 
  Moreover, the rankings do \emph{not} mean that a higher ranked strategy was \emph{always} better than a lower ranked one. The idea behind this representation is 
to give the user of a diagnosis system a guidance how to set parameters (diagnosis computation algorithm, number of computed diagnoses, heuristic used for measurement selection) in order to have the highest chance of achieving best estimations (EXP1) / efficiency (EXP2).   

\emph{Table~\ref{tab:best_wrt_seq-diag-time}} lists the best (ranked)
sample types 
wrt.\
overall time per diagnosis session in EXP2 (cumulated system computation time plus cumulated time for all measurements) for different scenarios and assumptions (1min, 10min) of measurement conduction times. The two rightmost columns (``adj'') show hypothetical results under the assumption that sample types \ref{T1} ($\rd$) and \ref{T6} ($\wf$)---which we naively simulated by means of brute force diagnosis computation in our experiments (cf.\ Sec.~\ref{sec:sampling_techniques})---were as efficiently computable as sample type \ref{T5} ($\bfs$). This allows to assess the added value of, e.g., \emph{efficient} random diagnosis sampling techniques.  

\noindent\textbf{Discussion.} We address each research question 
in turn: \vspace{3pt}

\noindent\ref{rq1}:\footnote{Remarks wrt.\ \ref{rq1}: \emph{(1)}~We had to leave out the $k=2$ scenarios as there were too few informative MPs 
which made these scenarios not reliably analyzable. \emph{(2)}~Values and rankings for other types of correlation coefficients (i.e., Spearman and Kendall) were very similar to the presented (Pearson) results. \emph{(3)}~Most correlation coefficients were statistically significant ($\alpha = 0.05$), except for a few $k = 6$ scenarios and some scattered $k = 10$ cases.}
 \emph{(Elimination rate, criterion E, Table~\ref{tab:rankings})} We see 
that $\rd$ is the sample type of choice, as one would expect. In numbers, the median correlation coefficients over all cases per scenario for (best,worst) sample type for $k\in\{6,10,20,50\}$ were 
$\{(0.76,0.5)$, $(0.83,0.52),$ $(0.95,0.7),$ $(0.98,0.85)\}$, which reveals that estimations were altogether pretty good for all sampling techniques. However, for $k \geq 20$, coefficients for $\rd$ manifested a significantly lower variance than in case of all other techniques, i.e., all coefficients for $\rd$ concentrated in the interval [0.9,1], whereas lowest coefficients for all other techniques lay between less than 0.5 and 0.7. 
Moreover, it stands out that $\wf$ allowed 
almost as accurate estimations as $\rd$. 
A possible explanation for these favorable results of $\wf$ is that there is usually a 
large number of minimal diagnoses with a very small probability, which is why the ``sub-population'' from which the $\wf$ diagnoses are ``selected'' tends to be larger (and thus more representative) than for other sample types, except for $\rd$ (where diagnoses are drawn at random from the \emph{full} population).
Finally, it is interesting that approximate methods ($\awf,\ard,\abf$) produced less representative samples than exact ones. And, although $\rd$ comes out on top for E, its approximate counterpart $\ard$ shows the worst results. \vspace{1pt}

\noindent\emph{(Probability, criterion P, Table~\ref{tab:rankings})} Here, $\bfs$ proved to be the predominantly superior technique in all depicted scenarios. Although the fact that $\rd$ was only the second best method might be surprising at first sight, the likely explanation for this is that often few of the most probable diagnoses already account for a major part of the overall probability mass, which is why they are more reliable for estimations of P than a random sample. For the same reason, it comes as no revelation that $\wf$ samples turned out to be the least preferable means to estimate P. The medians of the correlation coefficients over all cases per scenario for (best,worst) sample type for $k\in\{6,10,20,50\}$ were 
$\{(0.93,0.6)$, $(0.87,0.64),$ $(0.98,0.74),$ $(0.99,0.86)\}$. Thus, again, all sampling methods enabled pretty decent estimations, even for small sample sizes of only six diagnoses. Similarly as for E, the variance of correlation coefficients was significantly lower for the best sample type, $\bfs$, than for all others. However, the spread of correlation coefficients for the single sample types was noticeably larger for P than for E for $k \leq 20$, suggesting that only $\bfs$ facilitates very reliable estimations of P for small to medium sample sizes.\vspace{3pt}

\noindent\ref{rq2}: \emph{(Number of measurements, criterion M, Table~\ref{tab:rankings})} We find that $\bfs$ was the best strategy if all data is considered; and it was the most suitable choice for heuristics ENT and SPL and for small sample sizes $\{2,6\}$. On the other hand, it was the worst choice for the MPS heuristic where it led to substantial overheads (of up to $>$100\,\%) compared to other sample types, especially for large sample sizes. E.g., for the diagnosis problem U and $k = 50$, a diagnosis session using $\bfs$ involved 58 measurements vs.\ 25 measurements if $\rd$ was used instead. 
What is somewhat surprising is that $\bfs$ decidedly outperformed $\rd$ in the SPL scenarios, although the SPL function does not use any probabilities (where $\bfs$ leads to better estimations), but solely the elimination rate (where $\rd$ produces better estimates). Further analyses are needed to better understand this phenomenon.
Overall, $\rd$ compares favorably only against $\awf$ and $\wf$, but its performance depends largely on the used heuristic. For RIO it is even the sample type of choice, and for MPS it 
clearly overcomes $\bfs$. 
For all four heuristics, one of the approximate methods was the second best method, among which $\ard$ led to good performance most consistently. In comparison with $\rd$, $\ard$ was only (slightly) outweighed for RIO, but prevails for the other three heuristics. When considering large samples (20 or 50 diagnoses), $\ard$ even turned out to be the overall winner.  
This indicates that the QuickXPlain-based approximate random algorithm, in spite of its rather poor estimations (cf.\ E and P in Table~\ref{tab:rankings}), tends to be no less effective than a real random strategy. Finally, observe that $\wf$ was in fact the least favorable option in quasi all scenarios. \vspace{1pt}

\noindent\emph{(Time for diagnosis session, criterion T, Table~\ref{tab:rankings})}
Due to the naive brute force approach we used in our experiments to generate samples of type $\rd$ and $\wf$, it comes as no surprise that these two methods perform most poorly in terms of T. When drawing our attention to the best strategies, we find that, in all but one ($h = \text{SPL}$) of the shown scenarios, it is a different sample type that exhibited lowest time (T) than the one that manifested the lowest number of measurements (M). This prompts the conjecture of some time-information trade-off in diagnosis sampling---or: whenever the sampling process is most efficient (on avg.), the measurements arising from the sample are not most effective (on avg.). 
In particular, we recognize that, if an exact method is best for T (M), then an approximate method is best for M (T). And, unlike for M, $\ard$ tends to be worse than $\awf$ and $\abf$ in case of T.\vspace{1pt}   

\begin{table}[!t]
	\scriptsize
	\centering
	\caption{Best sample types wrt.\ overall sequential diagnosis time for various scenarios (EXP2).}
	\label{tab:best_wrt_seq-diag-time}
	\begin{tabular}{@{}l|cccc@{}}
		\toprule
		& \multicolumn{4}{c}{\textbf{best sample type} (given that time for each measurement $= t$)} \\ \midrule
		\textbf{scenario} & $t=1$min       & $t=10$min       & $t=1$min (adj)       & $t=10$min (adj)      \\ \midrule
		all data                     & $\bfs$         & $\bfs$          & $\bfs$               & $\bfs$               \\
		$k=2$                        & $\bfs$         & $\bfs$          & $\bfs$               & $\bfs$               \\
		$k=6$                        & $\bfs$         & $\bfs$          & $\bfs$               & $\bfs$               \\
		$k=10$                       & $\abf$         & $\abf$          & $\abf$               & ($\rd$,$\abf$)       \\
		$k=20$                       & $\awf$         & $\bfs$          & $\awf$               & $\bfs$               \\
		$k=50$                       & $\bfs$         & $\bfs$          & $\rd$                & $\rd$                \\
		$h=\text{ENT}$               & $\bfs$         & $\bfs$          & $\bfs$               & $\bfs$               \\
		$h=\text{SPL}$               & $\bfs$         & $\bfs$          & $\bfs$               & $\bfs$               \\
		$h=\text{RIO}$               & $\bfs$         & $\ard$          & $\rd$                & ($\ard$,$\bfs$)      \\
		$h=\text{MPS}$               & $\ard$         & $\ard$          & $\ard$               & $\ard$               \\ \bottomrule
	\end{tabular}
\end{table}

\noindent\emph{(Diagnosis session time, criteria T, M combined, Table~\ref{tab:best_wrt_seq-diag-time})}
Since the outcome for \ref{rq2} is not at all clear-cut when viewing M and T separately, we investigate their combined effect, i.e., the overall (avg.) length of diagnosis sessions for the different sample types. In brief, the conclusions are:
\begin{itemize*}[noitemsep]
	\item For small sample size below 10, go with $\bfs$. 
	\item For sample size 10, use $\abf$.
	\item For sample size 20, take $\awf$ if the expected time for conducting measurements is low, and take $\bfs$ else. 
	\item Unless there is an efficient method for $\rd$, use $\bfs$ for large sample size (50), otherwise use $\rd$.
	\item For ENT or SPL, adopt $\bfs$.
	\item For RIO, if measurement time is short, use $\bfs$ if there is no efficient algorithm for $\rd$, otherwise use $\rd$; if measuring takes longer, use $\ard$.
	\item For MPS, use $\ard$.
\end{itemize*}  \vspace{3pt}

\noindent\ref{rq3}: For theoretical representativeness, we observe pretty consistent (ranking) results over all sample sizes (cf.\ often equal entries in each column for each of the E and P criteria in Table~\ref{tab:rankings}). There is more variation when comparing results for different diagnosis problems. Nevertheless, results are fairly stable concerning the winning strategy: $\rd$ is in all cases the best (75\,\%) or second best (25\,\%) sample type for E, and $\bfs$ is in all but one case the best (63\,\%) or second best (25\,\%) sample type for P.
For practical representativeness, we see more of a fluctuation over different sample sizes and heuristics, as discussed for \ref{rq2} above (cf.\ variation over entries of each column for M and T in Table~\ref{tab:rankings}). Examining (ranking) results over different diagnosis problems reveals a similar picture, where however the rankings for T are decidedly more stable than those for M, meaning that relative sampling times are less affected by the particular problem instance than the informativeness of the samples.  \vspace{3pt} 

\noindent\ref{rq4}: Our data indicates a clear trend that increasing sample size leads to better theoretical representativeness (cf.\ discussion of \ref{rq1}). However, it also suggests that there is no general significant
positive effect of larger sample size on practical representativeness. While this is obvious for sampling time (T), i.e., generating more diagnoses cannot take less time, it is less so for the number of measurements (M). In fact, we even measured increases wrt.\ M in some cases (e.g., for MPS) as a result to drawing larger samples. 
This corroborates similar findings in this regard, albeit for lower sample sizes and other types of diagnosis problems, 
reported by \cite{rodler2018ruleML,dekleer1995trading}.\vspace{3pt} 
%

\noindent\ref{rq5}: From our data, we cannot generally conclude that a better theoretical implies a better practical representativeness (see discussion on \ref{rq1}, \ref{rq2} and \ref{rq3}). E.g., observe the performance of $\ard$ in the top vs.\ bottom part of Table~\ref{tab:rankings}. 
We surmise the cause of that to lie in the fact that \emph{(1)}~heuristics are based on a lookahead of one step only (where the approximate character of this analysis might counteract the benefit of good estimations), and \emph{(2)}~the added (information) value 
of additional diagnoses taken into a sample, regardless of how selected, decreases with the sample size (cf.\ the \emph{law of diminishing marginal utility} \cite{easterlin2005} in economics). 




\section{Research Limitations}
\label{sec:research_limitations}
Our evaluations do not come without limitations. In brief, there are the following threats to validity:\footnote{Those bullet points marked by a ``$*$'' we plan to address in terms of additional experiments as part of future work.}
\begin{itemize*}[noitemsep]
	\item For feasibility reasons, we \emph{(1)}~did not use \emph{all} diagnoses to determine the real values (EXP1), but just all minimal ones, and \emph{(2)}~used only such problem instances that reasonably allow the generation of all minimal diagnoses.
	\item We focused on binary-outcome measurements which are common in some, but not all diagnosis sub-domains, e.g., in ontology and KB debugging \cite{Rodler2015phd,rodler2019_expert_questions}, circuit diagnosis \cite{dekleer1987}, or matrix-based methods \cite{shakeri2000}.$*$
	\item We did not evaluate $\bfs$/$\wf$ sample types including minimum/maximum-cardinality minimal diagnoses, but concentrated on most/least probable ones.$*$
	\item To keep the size of our dataset manageable, we \emph{(1)}~omitted less commonly used existing heuristics, and \emph{(2)}~included only a 
	subset of available diagnosis computation methods into our analyses.$*$
	\item In EXP2, differences between sample types underlying the rankings were in many cases not statistically significant ($\alpha = 0.05$) due to the relatively low number of 10 diagnosis sessions we ran for each factor combination (note, EXP2 already takes several weeks of computation time with these settings). So, conclusions from our data must be treated with caution for the time being.$*$
\end{itemize*} 

\section{Conclusions}
\label{sec:conclusion}
We tested six diagnosis sampling techniques wrt.\ the quality of their estimations used by measurement selection heuristics, and 
their achieved performance in terms of diagnostic efficiency. Whereas random sampling, in line with statistical theory, leads to highly reliable estimations, this benefit is only conditionally reflected by the performance exhibited by random samples in diagnosis sessions, e.g., if the sample size is large 
or one specific heuristic is adopted. It turned out that, inspite of their missing statistical foundation, fully biased best-first samples including the most probable diagnoses were the best method in more of the investigated scenarios than any other sample type.
However, 
in the majority of scenarios some of the other sample types was better than best-first. This shows that a generally most favorable sampling technique cannot be nominated from our studies, and that the optimal sampling technique to draw on depends on the particular diagnosis scenario (e.g., sample size, measurement selection heuristic).   \vspace{3pt}

\noindent\textbf{Acknowledgments.} This work was supported by the Austrian Science Fund (FWF), contract \mbox{P-32445-N38}.

\fontsize { 9pt }{ 9pt } 
\selectfont

\end{document}